\documentclass[10pt,twocolumn]{article}

\usepackage{iccv}
\usepackage{graphicx}
\usepackage{amsmath}
\usepackage{amssymb}

 \iccvfinalcopy 

\ificcvfinal
\begin{document}

\title{Learning Hypergraph Labeling for Feature Matching}

\author{Toufiq Parag, Vladimir Pavlovic and Ahmed Elgammal\\
Dept of Computer Science, Rutgers University, NJ \\
{\tt \{tparag, vladimir,  elgammal\}@cs.rutgers.edu} }

\maketitle
\thispagestyle{empty}

\begin{abstract}

This study poses the feature correspondence problem as a hypergraph
node labeling problem. Candidate feature matches and their subsets
(usually of size larger than two) are considered to be the nodes and
hyperedges of a hypergraph. A hypergraph labeling algorithm, which
models the subset-wise interaction by an undirected graphical model,
is applied to label the nodes (feature correspondences) as correct
or incorrect. We describe a method to learn the cost function of
this labeling algorithm from labeled examples using a graphical
model training algorithm. The proposed feature matching algorithm is
different from the most of the existing learning point matching
methods in terms of the form of the objective function, the cost
function to be learned and the optimization method applied to
minimize it. The results on standard datasets demonstrate how
learning over a hypergraph improves the matching performance over
existing algorithms, notably one that also uses higher order
information without learning.
\end{abstract}

\section{Introduction}

Identifying feature correspondence is an important problem in
computer vision (see references in ~\cite{olivier-09}). In general,
matching features using only the appearance descriptor values can
often result in many incorrect matches. To address this problem,
most algorithms for feature correspondence combine information about
both appearance and geometric structure among the feature locations.
Several methods~\cite{berg05, Shapiro1992283,
leordeanu05,Cour:nips06, Gold96GA} utilize the pairwise geometric
consistency, along with the pointwise descriptor similarity, to
design a matching cost function which is minimized using various
optimization algorithms. For example,~\cite{Shapiro1992283,
Cour:nips06, leordeanu05} uses spectral techniques to compute a \lq
soft\rq~ assignment vector that is later discretized to produce the
correct assignment of features. These works model the appearance and
pairwise geometric similarity using a graph, either explicitly or
implicitly, and are commonly known as graph matching algorithms. The
soft assignment vector is typically computed by an
eigen-decomposition of the compatibility or the match quality
matrix. Several studies applied graph matching algorithms for
various vision problems~\cite{leordeanu09}.

Caetano et.al.~\cite{cateno09} discusses how the parameters of the
matching cost function (primarily the match compatibility scores)
can be learned from pairs with labeled correspondences to maximize
the matching accuracy.  A more recent work~\cite{leordeanu09}
proposes to learn similar matching scores in an unsupervised fashion
by repeatedly refining the soft assignment vector.

Higher order relationship among the feature points have also been
investigated as the means of improving the matching accuracy.  Zass
et.al.~\cite{zass08} assumes two separate hypergraphs among the
feature points on two images and propose an iterative algorithm to
match the the two hypergraphs. On the other hand, Olivier
et.al.~\cite{olivier-09} generalize the pairwise spectral graph
matching methods for higher order relationships among the point
matches. The pairwise score matrix is generalized to a high order
compatibility tensor. The eigenvectors of this tensor are used as
the soft assignment matrix to recover the matches.

In our framework, each feature correspondence is considered as a
datapoint and we assume a hypergraph structure among these
datapoints (similar to~\cite{olivier-09}). That is, we conceive a
subset of candidate feature matches as a hyperedge of the
hypergraph.  For subsets of such datapoints, we assume that the
relationship among features of one image follows the same
geometrical model as that present among the corresponding features
in the other image. We compute the likelihood, using this
geometrical model, for every subset of datapoints and use it as
weight of the hyperedge. The objective is to label the datapoints,
i.e., matches to be correct or incorrect, given this hypergraph
structure among them.

We adopt a hypergraph node labeling algorithm proposed
in~\cite{parag11}. Given a hypergraph, where the hyperedge weights
are computed using a model, this algorithm produces the optimal
labeling of the nodes that maximally conforms with the hyperedge
weights or likelihood values. Within the framework, the higher order
interaction among subsets of datapoints is modeled using a higher
order undirected graphical model or the Markov network
(see~\cite{parag11} for details). The labels are computed by solving
the inference problem on this graphical model where a labeling cost
or energy function is minimized to produce the optimal labeling.

In this paper, we show that the framework of hypergraph node
labeling of~\cite{parag11} can be applied for feature matching. In
addition, we show how it is possible, and in fact advantageous, to
learn (a parametric form of) the)cost function for matching given
several labeled examples of feature correspondences. The learned
forms of cost functions are able to appropriately weight the label
disagreement cost for different subsets. For example, if the number
of subsets containing more accurate matches than the inaccurate
ones, the associated penalty function will attain a higher weight to
balance the relative importance. The learning procedure is general,
i.e., in addition to the feature matching, it can be utilized for
any application of the labeling problem~\cite{parag11}.

%


Point matching problem was addressed by a probabilistic graphical
model before, in~\cite{caetano06graphical, Komodakis_Paragios_2008},
enforcing a graph among the points for spatial consistency. The
required potential (cost) functions in these two studies were
pre-selected and not learned from the data. Our approach can handle
match interaction in larger sets and demonstrates the advantage of
learning the cost functions from the data. Feature matching problem
has also been cast as an energy minimization problem
in~\cite{Torresani2008}.

\subsection{Contribution:} 
\begin{figure}[h]
\vspace{-0.2cm}
\begin{center}
\parbox{1.25in}{\centerline{\includegraphics[width=1.25in,height=1.25in]{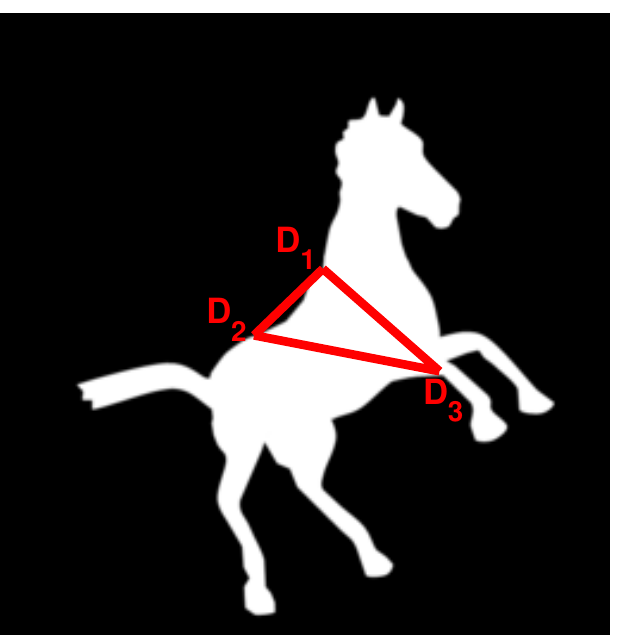}}}
\parbox{1.25in}{\centerline{\includegraphics[width=1.25in,height=1.25in]{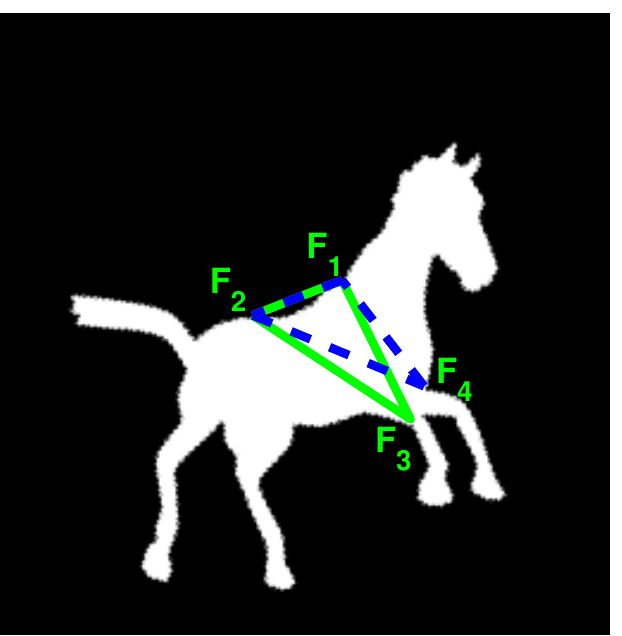}}}
\caption{\small Triangle pairs with overlapping matches.} \label{F:MOTIVATION}
\vspace{-0.5cm}
\end{center}
\end{figure}

At this point, we would like to clarify
what aspect of learning (hypergraph labeling for) point matching is
different from earlier works. Let us suppose $x_i \in \{ 0, 1\}$ is
the label for $i$-th candidate feature match, $x_i = 1$ implies a
correct match and $x_i = 0$ implies an incorrect one. Let $H_{V^k}$
be the match compatibility score of a subset $V^k$ of matches of
size $k$. The popular graph and tensor matching algorithms maximize
the following overall matching score to retrieve the correct
matches~\cite{leordeanu05, cateno09, Cour:nips06, olivier-09}.
\vspace{-0.2cm}
\begin{equation}\label{E:SCORE}
\small S(X) = \sum_{V^k} H_{V^k}~ \prod_{l=1}^k x_{i_l}.
\vspace{-0.2cm}
\end{equation}
The score function is a weighted summation of subset-wise label
concurrence function, $s(V^k) = \prod_{l=1}^k x_{i_l}$. Notice that,
$s(V^k)$ is a binary valued function: $s(V^k) = 1$ only when all
labels $x_{i_1}, \dots, x_{i_k}$ are equal to 1 and 0 otherwise.
Instead of using this predefined binary valued function, we
investigate whether or not such label agreement function (or,
conversely a disagreement cost function) can be learned from labeled
matches. We believe it is particularly useful to learn this function
for higher order ($k > 2$) methods. To illustrate the necessity of
such learning, we show two images in Figure~\ref{F:MOTIVATION} with
candidate feature matches $(D_1, F_1), (D_2, F_2), (D_3, F_3)$ and
$(D_3, F_4)$ , all with equal matching probability, overlaid on
them.

It is assumed that the geometrical arrangement among matching
features can be encoded by triangle. Clearly, the similarity between
triangles $D_1 D_2 D_3$ (red) and $F_1 F_2 F_3$ (green) will be high
resulting in a large match compatibility $H_{V^3}$ (where $V^3 = \{
(D_1, F_1), (D_2, F_2), (D_3, F_3) \}$). Notice that the triangle
$F_1 F_2 F_4$ (blue dashed) would also have relatively large
similarity with $D_1 D_2 D_3$. Though this subset $\{(D_1, F_1),
(D_2, F_2), (D_3, F_4) \}$ of matches contain one incorrect match
$(D_3, F_4)$, it still provides us significant geometric information
about the two correct matches $(D_1, F_1)$ and $(D_2, F_2)$ .
Incorporating this information in the algorithm should assist
establishing more correct correspondences among the features.
However, the form of $s(V^k) = \prod_{l=1}^k x_{i_l}$ does not
explicitly handle this situation, even when $x_{i_l}$ is relaxed to
take values in real domain\footnote{For binary $x_{i_l}$, $s(V^3) =
0 $ with one incorrect match in $V^3$ and therefore the
compatibility score is ignored.}. One needs to learn an appropriate
label agreement (or disagreement cost) function to explicitly
include this information in the framework. Learning the cost
function can also counteract the uneven ratio of subsets with more
correct matches and those with more incorrect matches.

As it will be explained in details later, to determine the
correspondence, we in fact \emph{minimize a cost function} of the
form as follows. \vspace{-0.2cm}{\small
\begin{align}
\tilde{{\cal E}}(X) = & \sum_{V^k} H_{V^k}~ \tilde{g}_1(x_{i_1}, \dots, x_{i_k}) \nonumber \\
 &+ (1- H_{V^k})~ \tilde{g}_0(x_{i_1}, \dots, x_{i_k}).
\vspace{-0.3cm}
\end{align}}
\noindent This paper describes how to learn appropriate subset-wise
label disagreement cost functions (also referred as penalty
functions) $\tilde{g}_1$ and $\tilde{g}_0$ from labeled matches. Our
approach is  significantly different in concept from previous
learning algorithms for correspondence. The algorithms
of~\cite{cateno09, leordeanu09} aim to learn a match compatibility
function $H_{V^k}$ from the data to optimally reflect accurate
correspondences among the features. On the contrary, our algorithm
learns the label disagreement cost functions $\tilde{g}_1$ and
$\tilde{g}_0$ to minimize the total label disagreements within the
subsets \emph{given the subset matching qualities $H_{V^k}$}.
 The next section describes how feature correspondence can
be cast as a hypergraph labeling problem as defined
in~\cite{parag11}



\section{Matching as hypergraph labeling}\label{S:MRFFORM}

Given two images $I_L$ and $I_R$, we denote $a_l$ and $a_r$ to be
the indices of  feature points from $I_L$ and $I_R$ respectively. In
general, the number $n_L$ of features in $I_L$ is different from the
number $n_R$ of features in $I_R$. Each candidate match $(a_l, a_r)$
is considered to be a datapoint $v_i,~i=1, \dots, n,$ in our
approach. The goal is to partition the dataset $V = \{ v_1, \dots,
v_n\}$ into subset $A$ comprising correct correspondences and to $B$
comprising incorrect ones. This is a data labeling problem where the
binary label $x_i \in\{0, 1\}$ of $v_i$ needs to be assigned $x_i =
1$ if $v_i$ belongs $A$ and to $0$ otherwise.

We wish to exploit the information about subsets of datapoints to
enforce geometric consistency in matching. More specifically, for a
subset $ V^k = \{ v_{i_1}, \dots, v_{i_k}\} = \{(a_{l_1}, a_{r_1}),
\dots, (a_{l_k}, a_{r_k}) \}$ of size $k$ of matching points, we
assume the geometric relationship among $\{a_{l_1}, \dots, a_{l_k}
\}$ to be similar to that among $\{a_{r_1}, \dots, a_{r_k}\}$. This
similarity value (computed by a suitable function) is denoted by
$\lambda(V^k) \in [ 0, 1 ]$. Notice that, we are effectively dealing
with a hypergraph with datapoints $v_i$ as the nodes and the subsets
$V^k$ as the hyperedges. Given such hypergraph, the labeling
algorithm is supposed to partition the set of nodes into two sets
$A$ and $B$, corresponding to correct and incorrect matches
respectively. We will use the term likelihood value and weight
interchangeably when referring to similarity value $\lambda(V^k)$.

The work in~\cite{parag11} models the higher order interactions in
this hypergraph by a Markov network (by a Conditional Random Field
(CRF) to be precise)~\cite{taskar07book}. The optimal labeling can
then be achieved by solving the inference for this CRF model. We
follow this representation which is described in the next section.

\section{The cost function}\label{S:PROPCLIQ}

Let ${\cal V}^k$ be the set of all hyperedges $V^k$ in this
hypergraph.  Let $X = \{ x_1, \dots , x_n \}$ be a label assignment
of the nodes $V$ of the hypergraph. The cost function that asserts
discrepancy of  node assignments $X$ in the hypergraph nodes $V$ can
be written as

\vspace{-0.2cm}
\begin{equation}
\small {\cal E}(X,V) = \sum_{V^k \in {\cal V}^k} E^k(X^k,V^k),
\vspace{-0.2cm}
\end{equation}

where $X^k$ is the set labels of member nodes of subset $V^k$ and
$E^k$ is the local discrepancy, i.e.,  the cost of assignment $X^k$
in $V^k$. We assume functionally homogeneous local costs, $E^k = E$.
Given this representation, it is possible to construct an equivalent
CRF with clique potentials $E^k$ (see~\cite{parag11}) and formulate
the optimal assignment task as the inference in this CRF.

Following~\cite{parag11} , each clique potential $E$ is represented
as

\vspace{-0.2cm}
\begin{equation} \label{E:SUBCLIQ}
\small E(X^k ; V^k) = \ \beta_1 \ \lambda(V^k)\ g_1(\eta_0) \ +\ \beta_0\
(1-\lambda(V^k)) \ g_0(\eta_1) .
\vspace{-0.2cm}
\end{equation}

Here, $g_c,~c = 0, 1$ represent a penalty function : the cost of
assigning clique nodes to an incorrect class (eg, match to non-match
and vice-versa).  The penalty function is defined as a function of
$\eta_{1-c}$, the number of nodes in the clique whose label differs
from the clique hypothesis $c$.  $\beta_c$ are non-negative
balancing parameters and $\eta_0 + \eta_1 = k$.

Intuitively,  this potential penalizes, via functions $g_c$, the
label assignments incompatible  with one of the two hypotheses,
matching and non-matching features. To achieve this, the penalties
$g_c$ should be non-decreasing in $\eta_{1-c}$. If the likelihood of
matching, $\lambda(V^k)$, is high, the potential seeks to decrease
$\eta_0$, the number of assignments to "not-matching" hypothesis. In
the opposite case, with high non-matching likelihood
$1-\lambda(V^k)$, the potential attempts to decrease the number of
labels incompatible with this hypothesis, $\eta_1$.

Penalty functions $g_c$ could be directly modeled as linear and
nonlinear functions of number of label disagreement $\eta_{1-c}$ in
the clique. However, as it will become clear later, it is
advantageous to learn a nonlinear mappings $g_c$ from labeled data.
The next section describes how the functions $g_c$ can be learned
from labeled matches/mismatches.

\section{Learning penalty functions}\label{S:LEARN}

Given $J$ hypergraphs with hyperedges ${\cal V}^k_j,~j = 1, \dots,
J,$ along with the weights and labels $X_j$ of the datapoints (or
correspondences), we wish to learn the parametric form of the $g_c$
functions. We first describe two parametric forms of the penalty
functions so that the clique potentials, as defined in
Equation~\ref{E:SUBCLIQ} become log-linear models. In particular, we
seek to express the potential as a linear combination of factors
defined over each clique)~\cite{taskar07book} \vspace{-0.2cm}
\begin{equation}
\small E(X^k ~;~V^k) = \sum_l w_l \phi_l(X^k ; V^k).
\vspace{-0.2cm}
\end{equation}

In this definition, $\phi_l(X^k ; V^k)$ are the factors and $w_l$
are the mixing weights. The following sections explain how restating
the penalty functions in this manner facilitates learning using CRF
training algorithms.

\subsection{Discrete $g_c$}\label{S:LEARN_DISCRETE}
First, we express $g_c$ as a discrete function. Observe that,
penalty functions $g_c$ are defined on $\eta_{(1-c)}$ values, which
are integers in our case. Therefore, it suffices to learn a set of
discrete mapping $g_c(\eta_{(1-c)})$ for all $c \in \{0, 1\}$ and $0
\le \eta_{(1-c)} \le k$. Let us introduce two quantities as follows
\vspace{-0.2cm}{\small
\begin{align}\label{E:LEARN_DISCRETE}
\small w_c^\alpha &= \beta_c g_c(\alpha), \\  \phi_c(\alpha ; V^k) &= -
\lambda_c(V^k)~I(\eta_{(1-c)}, \alpha),
\vspace{-0.2cm}
\end{align}}
\noindent where $I(s , t)$ is an indicator function which equals to
$1$ only when $s$ is equal to $t$ and $0$ otherwise. Furthermore,
the likelihood weights are denoted by $\lambda_1(V^k) =
\lambda(V^k)$ and $\lambda_0(V^k) = 1 - \lambda(V^k)$ for notational
convenience. Notice that, in this case, $\phi_c$ functions are the
factors (for each clique) that assume nonzero values only when
$\eta_{1-c} = \alpha$. The clique cost function defined in
Equation~\ref{E:SUBCLIQ} can be rewritten as follows \vspace{-0.2cm}
\begin{equation}\label{E:POTEN_DISCRETE}
\small E(X^k ; V^k) = \sum_{c} \sum_{\alpha=0}^k w_c^\alpha \phi_c(\alpha;
V^k).
\vspace{-0.2cm}
\end{equation}

This definition of $g_c$ expresses the joint probability of any
assignment as log-linear model. For this form of $g_c$, the values
of $w_c^\alpha$ are learned for all $\alpha = 1, \dots,
\eta_{(1-c)}$ and $c = 0, 1$.

\subsection{Second order polynomial $g_c$}\label{S:LEARN_DERIV}
Unconstrained forms of $g_c$ may be prone to overfitting.  We thus
propose a more constrained $g_c$ by assuming a second order
polynomial form for it. In this case, this function can be expressed
using the Taylor expansion around reference point $0$:
\vspace{-0.2cm}
\begin{equation}\label{E:TAYLOR}
\small g_c(\alpha) = g^{(0)}_c + \alpha g^{(1)}_c + {\alpha^2 \over 2}
g^{(2)}_c.
\vspace{-0.2cm}
\end{equation}
\noindent In Equation~\ref{E:TAYLOR}, $g^{(0)}_c, g^{(1)}_c$ and
$g^{(2)}_c$ are the 0, 1st and 2nd order derivatives of $g_c$ at
$0$. The features for this case can be defined as \vspace{-0.2cm}
{\small
\begin{align}\label{E:LEARN_POLY2}
& \psi^0_c(\alpha ; V^k) = -~\sum_{\gamma=0}^k \lambda_c(V^k)~I(\eta_{(1-c)}, \gamma),  \\
&  \psi^1_c(\alpha ; V^k) = -~\sum_{\gamma=1}^k \alpha~ \lambda_c(V^k)~
I(\eta_{(1-c)}, \gamma), \\  & \psi^2_c(\alpha ; V^k) = -~\sum_{\gamma=1}^k
{\alpha^2 \over 2} ~\lambda_c(V^k) ~I(\eta_{(1-c)}, \gamma).
\vspace{-0.2cm}
\end{align}}

Then, the cost function in Equation~\ref{E:SUBCLIQ} can be expressed
as linear combination of features $\psi^e_c(\alpha), ~ e= 0, \dots,
2$

\vspace{-0.2cm}
\begin{equation}\label{E:POTEN_POLY}
\small E(X^k ; V^k) = \sum_{c} \sum_{e=0}^2 g^{(e)}_c \psi^e_c(\alpha ; V^k).
\vspace{-0.2cm}
\end{equation}

For polynomial $g_c$, we learn the values of $g^{(e)}_c$ for all $e
= 0, 1, 2$ and $c = 0, 1$. This redefinition of $g_c$ has the
benefit of regulating the learned form to be of some specific type.
Also, regardless of the size $k$ or data subset, we only need to
learn $3\times C$ parameters, where $C$ is the total number of
classes. Next section briefly discusses existing techniques for
learning CRFs.

\subsection{Learning algorithms}\label{S:LEARN_OPT}
In last two sections we have shown that the clique potential
function of the proposed framework can be expressed as a linear
combination of features or factors. The joint probability of any
label configuration for a CRF, with discrete form of $g_c$, can be
stated as follows

\vspace{-0.2cm}
\begin{equation}\label{E:PX}
\small p(X~|~V) = {1 \over Z(V)}~ \exp \biggl \{\sum_{V^k \in {\cal V}^k}
\sum_{c=0}^1 \sum_{\alpha=0}^k w_c^\alpha ~\phi_c(\alpha , V^k) \biggr \}
\vspace{-0.2cm}
\end{equation}

\noindent where $Z(V)$ is a normalizing term, $Z(V) = \sum_X
p(X~|~V)$. The joint probability will be similar for second order
polynomial $g_c$ and we are omitting the derivation for it here.
There are two types of algorithms to estimate the parameters
$w_c^\alpha$ from data: one that aims at determining the parameters
by maximizing the log-likelihood~\cite{taskar07book} and the other
that maximizes the separation, or the label margin, between classes
of datapoints~\cite{Bartlett05EG}.

\subsubsection{Likelihood Maximization}

The log-likelihood function for the training data is given by
\vspace{-0.2cm}
\begin{equation}\label{E:LIKELIHOOD}
\small l(w) = \sum_{j=1}^J \sum_{V^k \in {\cal V}_j^k} \sum_{c=0}^1
\sum_{\alpha=0}^k w_c^\alpha ~\phi_c(\alpha ; V^k) - \log~Z(V).
\vspace{-0.2cm}
\end{equation}
It has been shown that $l(w)$ is concave~\cite{taskar07book}.
Therefore, a Gradient Ascent algorithm is able to produce the
globally optimal values for $w_c^\alpha$. It is straightforward to
see that the gradient with respect to $w_c^\alpha$ is the difference
between summation of observed and expected $\phi_c(\alpha)$ values
\vspace{-0.2cm}{\small
\begin{align}\label{E:GRADIENT}
{\partial l \over \partial w_c^\alpha} &= \sum_{j=1}^J \sum_{V^k \in
{\cal V}_j^k} \phi_c(\alpha ; V^k) \nonumber \\ &- \sum_{j=1}^J \sum_{V^k \in {\cal
V}_j^k} \sum_{X^k} \phi_c(\alpha ; V^k)~p(X^k ~|~ V^k ).
\vspace{-0.2cm}
\end{align}}
We used a sum-product belief propagation
algorithm~\cite{Kschischang98factorgraphs} to compute the marginal
posteriors $p(X^k ~|~ V^k)$. A regularizer term was added to the
likelihood function to penalize large parameter values. Apart from
Gradient Ascent, other algorithms such as Conjugate Gradient and
L-BFGS have also been for this maximization
problem~\cite{taskar07book}.

\subsubsection{Margin maximization}

The second type of algorithms try to estimate the parameters by
maximizing the class margin of the labeled examples. Margin
maximization is useful if the data distribution is biased to one of
the classes or there are many noisy samples in the data. Bartlett
et.al.~\cite{Bartlett05EG} proposed a constrained optimization
problem, in terms of primal variables $w_c^\alpha$, for parameter
learning in maximal margin setting. Their formulation minimizes a
loss function, defined in terms of the number of incorrectly labeled
examples, and a regularizer term. An exponentiated gradient (EG)
algorithm is applied to minimize the objective that updates the
primal variables $w_c^\alpha$ similarly as in
Equation~\ref{E:GRADIENT}. In addition, the EG algorithm also
updates the the dual variables  to minimize the subset-wise
mislabeling error. Furthermore, the marginal terms are different
from those in likelihood maximization -- in~\cite{Bartlett05EG},
they are calculated from a Markov network where the dual variables
act as potential functions.

More efficient version of both these algorithms have been described
in~\cite{collins-08}. In our experiments, parameters were learned by
standard Gradient Ascent optimization to maximize the likelihood for
a discrete $g_c$.

\section{Inference}\label{S:INF}

Once $g_c(\cdot),~ c \in \{0, 1\}$, are learned, problems with
nonlinear $g_c(\cdot)$ can be solved using any efficient Markov
network inference algorithm, See~\cite{ishikawa-09cvpr, werner-07},
and references therein. We adopted the sum-product belief
propagation~\cite{Kschischang98factorgraphs} since we also use it
for computing the marginal probabilities $p(X^k~|V^k)$ required to
learn the parameters. The output of this algorithm is belief
(approximate marginal probability)  $b_i(1)$ and $b_i(0)$ that any
datapoint $v_i$ belong to class $1$ and $0$ respectively.

The belief values for each datapoint $v_i$ could be used to
determine the hard one to one assignment for any feature $a_l$ of
image $I_L$ to its unique match $a_r$ on image $I_R$. To do this,
for each $a_l$, we select the match corresponding to the datapoint
with the largest ratio of two beliefs ${b_i(1) \over b_i(0)}$ among
all the datapoints associated with $a_l$. The accompanying feature
$a_r$ on the right image is selected as the resultant match for
$a_l$. This method of discretization is similar
to~\cite{leordeanu05}.


\section{Experiments and Results}\label{S:RESULT}
This section describes different matching experiments conducted on
standard datasets to test the proposed method and compares the
performances with past studies. For all the experiments, the penalty
functions were learned using Gradient Ascent to maximize the
likelihood for a discrete mapping $g_c$
(Section~\ref{S:LEARN_DISCRETE}).

\begin{figure*}[t]
\vspace{-0.6cm}
\begin{center}
\parbox{1.65in}{\centerline{\includegraphics[width=1.65in,height=1.5in]{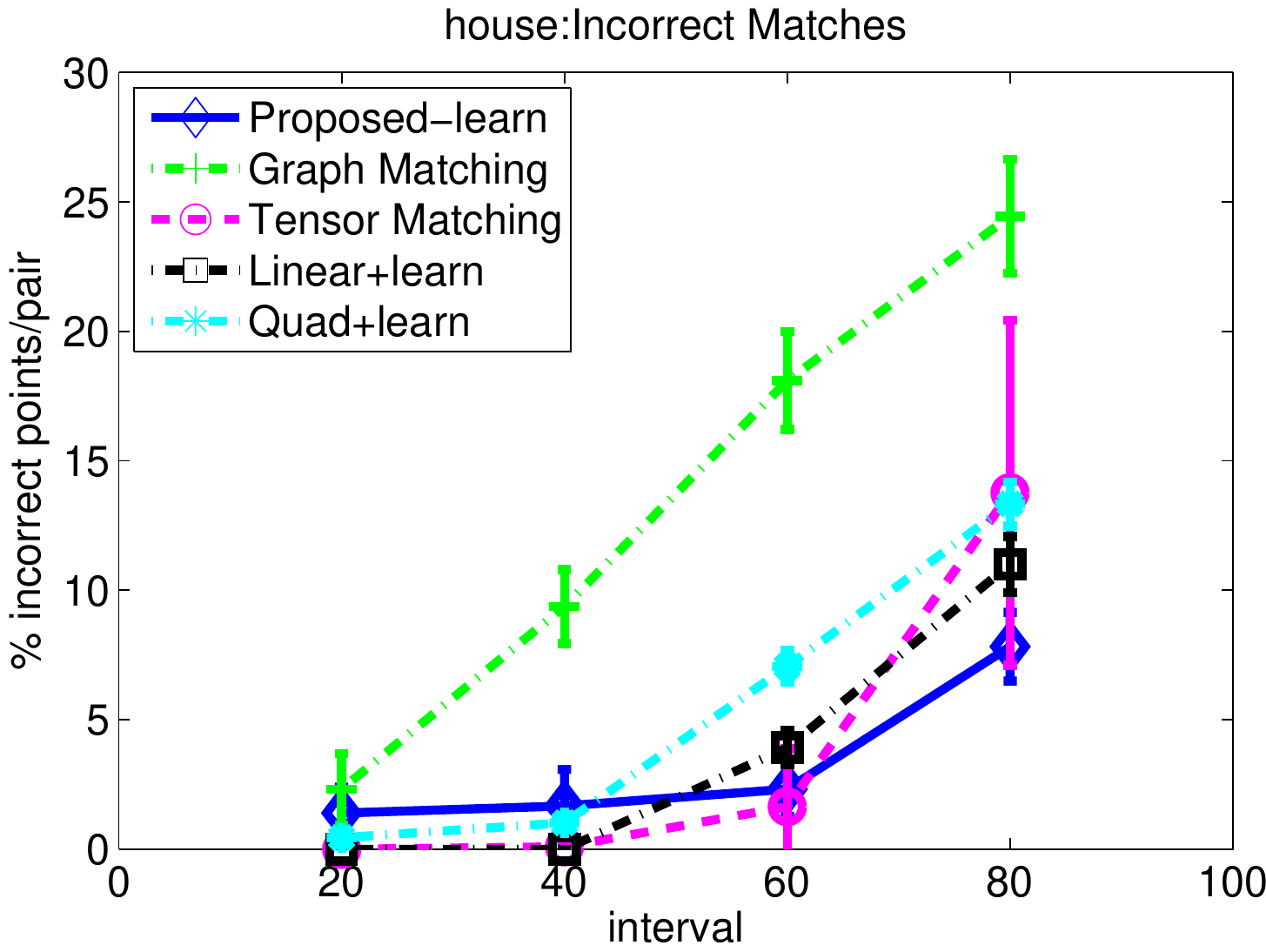}} }
\parbox{1.65in}{\centerline{\includegraphics[width=1.65in,height=1.5in]{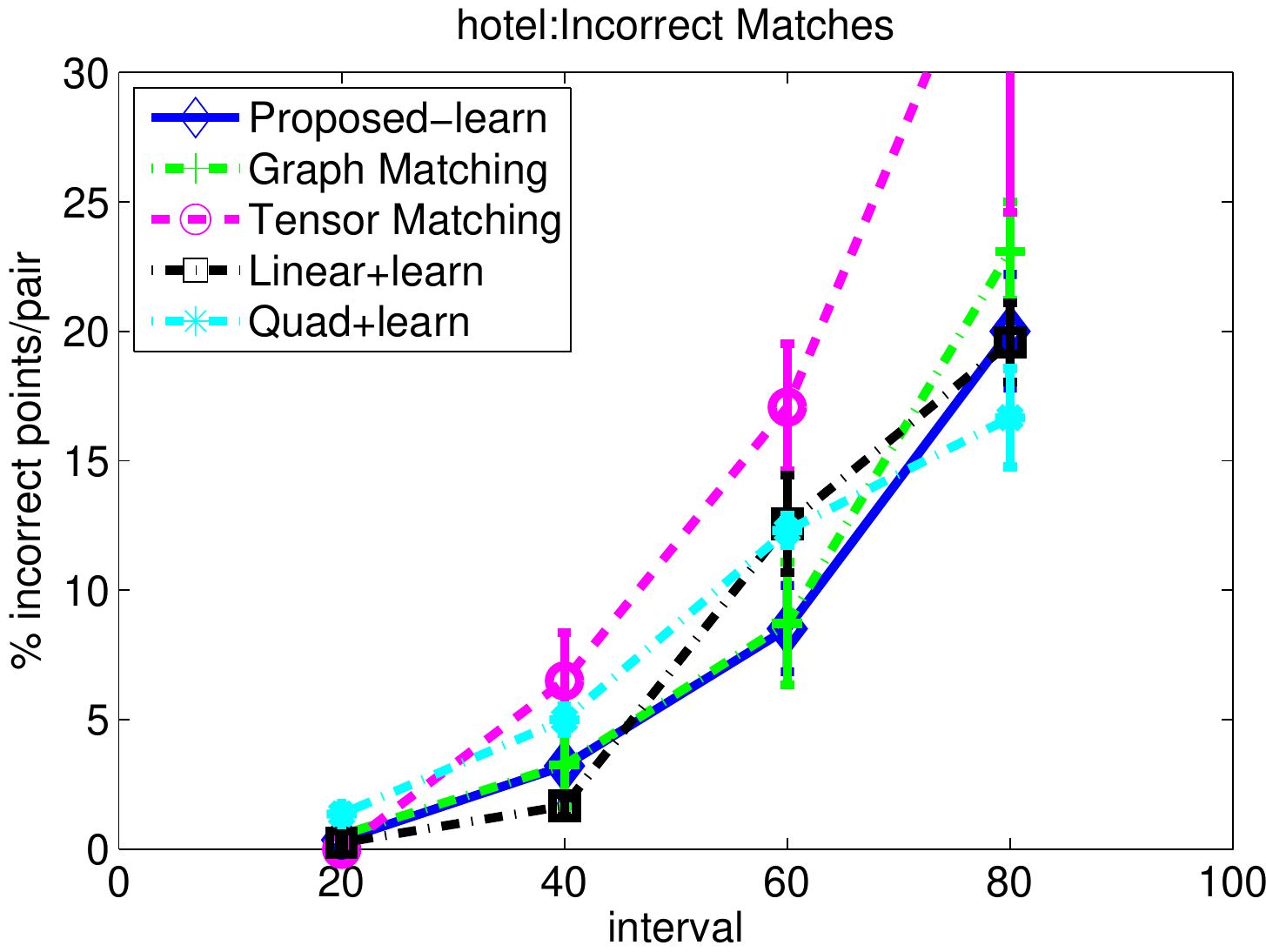}} }
\parbox{1.65in}{\centerline{\includegraphics[width=1.65in,height=1.5in]{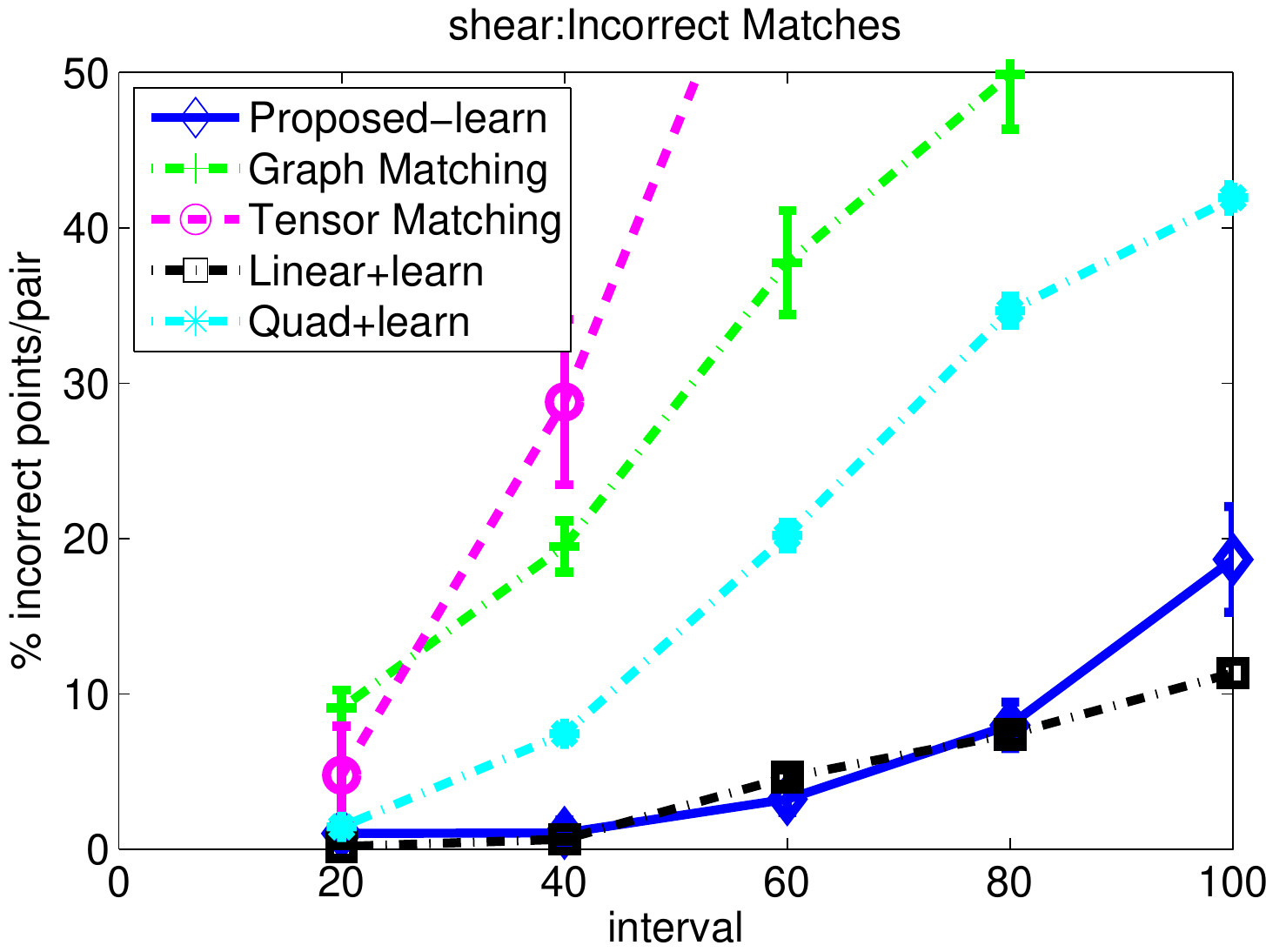}} }
\parbox{1.65in}{\centerline{\includegraphics[width=1.65in,height=1.5in]{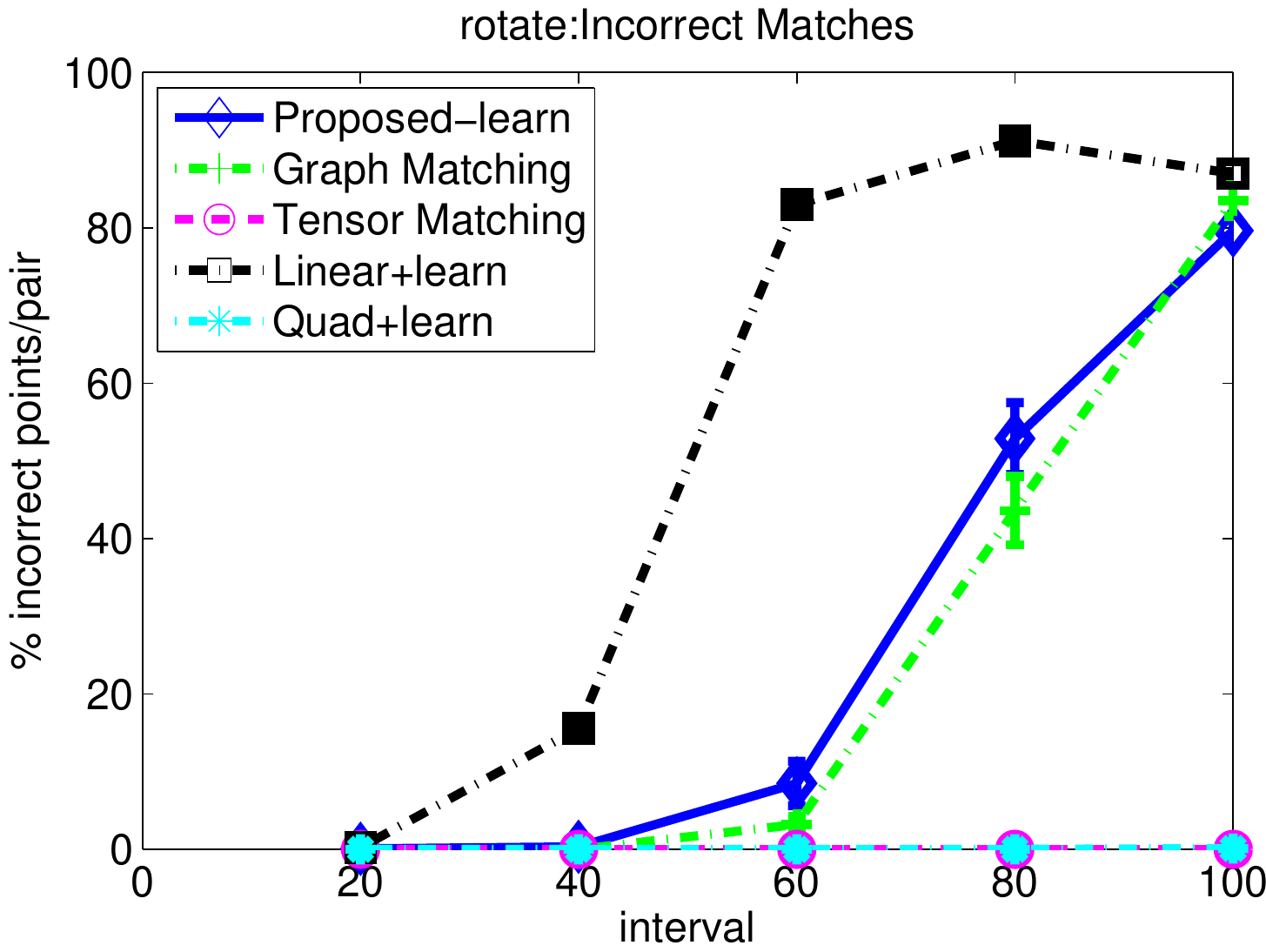}} }
\caption{\small (Left to right) House, Hotel, Horse-Shear,
Horse-Rotate: Mean and std deviation of incorrect matches.}
\label{F:QUANHOUSE} \vspace{-0.5cm}
\end{center}
\end{figure*}

\subsection{House, Hotel and Horse data}

We conduct our first experiment on the standard House and Hotel
datasets. Each of these datasets contains a sequence of (around 100)
images of a toy house (or hotel) seen from increasingly varying
viewpoint. Locations of a set of keypoints, that appear on each of
the image of the sequence, are available for both these sequences.

Another synthetic dataset, namely the silhouette  images of a Horse
as used in~\cite{cateno09}, were also included in this experiment.
From a single silhouette image, two sequences of 200 images were
generated by shearing and rotating. The width of the image is
sheared to twice of its height at most and the maximum angle of
rotation was 90 degrees. These image transformations are different
from those present in House and Hotel datasets. The feature
locations are extracted by a sampling method as in~\cite{cateno09}.

For the proposed algorithm, the Geometric Blur (GB)~\cite{berg01gb}
descriptor is used to represent each feature. For each keypoint
$a_l$ in image $I_L$, $m=3$ candidate matches, denoted by the set
$\mu(a_l)$, are chosen based on largest normalized correlation
between the GB descriptors. Each of the candidate matches is
considered to be a datapoint $v_i$.

We construct a hypergraph of edge cardinality $k=3$ with these
datapoints. For each feature point $a_l$ in image $I_L$, all
possible triangles are generated among $a_l$ and $k_{NN} = 5$
nearest neighbors. Any such triangle among $\{ a_{l_1}, \dots,
a_{l_k}\}$, has $k^m$ possible matching triangles in image $I_R$
induced by the set of candidate matches  $\{\mu(a_{l_1}), \dots,
\mu(a_{l_k})\}$. This construction of hypergraphs among matches
follows that of~\cite{olivier-09} and~\cite{zass08},
except~\cite{olivier-09} searches all possible triangles in image
$I_R$ instead of searching the ones induced by candidate matches.
The geometric similarity of these triangle pairs are evaluated by
the sum of squared difference of the angles similar to the tensor
matching algorithm~\cite{olivier-09}. The parametric difference
$\epsilon$ between triangles is converted to geometric similarity
weight using $1 - {\epsilon \over \delta }$ where $\delta =0.5$ for
all experiments in this section\footnote{Triangle pairs with
$\epsilon > \delta$ are discarded.}.



The appearance similarity value is the normalized correlation
between two GB descriptors computed for potential matching features.
Each candidate match is assigned a weight that reflects the quality
of the match computed by normalized correlation~\cite{leordeanu05}.
To compute the overall similarity $\lambda_1(V^k)$ between two
triangles, the weight of corresponding matches $\{\mu(a_{l_1}),
\dots, \mu(a_{l_k})\}$ is multiplied with the geometric similarity
weight computed from parametric difference between two triangles.


We consider four sets of image pairs where, in each pair, the two
images are $\{20, 40, 60, 80\}$ frames apart from the other (also
$100$ for Horse datasets). For each set of image pairs, first five
pairs were selected to learn the parameters for the proposed
matching algorithm. We learned the parameters for a discrete $g_c$
by maximum likelihood (ML) method (refer to Section~\ref{S:LEARN}).

The performance of our algorithm is compared against the following
algorithms:
\begin{enumerate}
  \item Tensor matching method~\cite{olivier-09}(implementation available at author's
website): The parameter values such as number of triangles to be
generated, number of nearest neighbors of each triangle and the
distances are tuned to produce the best results in each of the
experiments.
  \item Graph matching of~\cite{leordeanu05}: We used the exact same procedure
as described in the paper with the same $m = 3$ candidate
matches for each keypoint and the used $3$ as the distance
threshold to determine the neighboring keypoints (also tuned for
best result).
  \item Learning graph matching~\cite{cateno09}: The results of
  learning both the linear and quadratic assignments have been
  used for comparison.
\end{enumerate}
Figure~\ref{F:QUANHOUSE} shows the percentage of incorrect matches
produced by these and proposed method. Some qualitative results are
supplied as supplementary material.

%
%

The results show that none of the spectral Graph matching and Tensor
matching techniques was able to perform well on all of these
datasets. On the other hand, the proposed method, with learned cost
functions is more robust and accurate than all other methods in
House, Hotel and Horse-shear datasets. The result of learned Linear
Assignment procedure of~\cite{cateno09} closely follows that of our
method. However, learning linear assignment produces unacceptably
high error rates (much higher than the proposed method) for
Horse-rotate dataset. This is due to the fact that Linear Assignment
learns the weight vector for descriptor similarity for a candidate
match. Unless the window-- in which the descriptor is computed-- is
also rotated, the descriptor similarity would be too low in rotated
images for a weight vector to generate a correct match. This
observation supports the claim made in~\cite{leordeanu09} that, in
general,  Linear Assignment alone can not result in accurate
matches. The proposed algorithm and Graph
matching~\cite{leordeanu05} could not identify the correct matches
for larger rotational angles ($>$80 degrees) due to inferior initial
candidate matches.

These results attest the advantage of using higher order information
and learning the cost function for matching. Utilizing higher order
information consistently produced higher accuracy than learning
Quadratic Assignment in all but one dataset. The Tensor matching
algorithm, which uses higher order information but does not lear
from data,  was not robust either on different
datasets\footnote{In~\cite{olivier-09}, the authors did not report
the results on all possible pairs of images. Results for one pair of
images for each interval on House dataset were reported are these
values are the same as the minimum error rates of our result.}. The
reason for this behavior was surmised in the introduction: the
number of subsets generated by higher order algorithm is usually
large with imbalanced ratio of useful subsets. One needs to learn
the appropriate cost functions for accurate labeling of the members
of these subsets. However, it is interesting to see that both
Quadratic Assignment~\cite{cateno09} and Tensor
matching~\cite{olivier-09} produced a perfect matching for rotated
images (Figure~\ref{F:QUANHOUSE}, rightmost plot).
Indeed,~\cite{olivier-09} also reports similar matching results on
synthetic 2D points.


\begin{figure}[h]
\vspace{-0.3cm}
\begin{center}
\parbox{1.25in}{\centerline{\includegraphics[width=1.25in,height=1.25in]{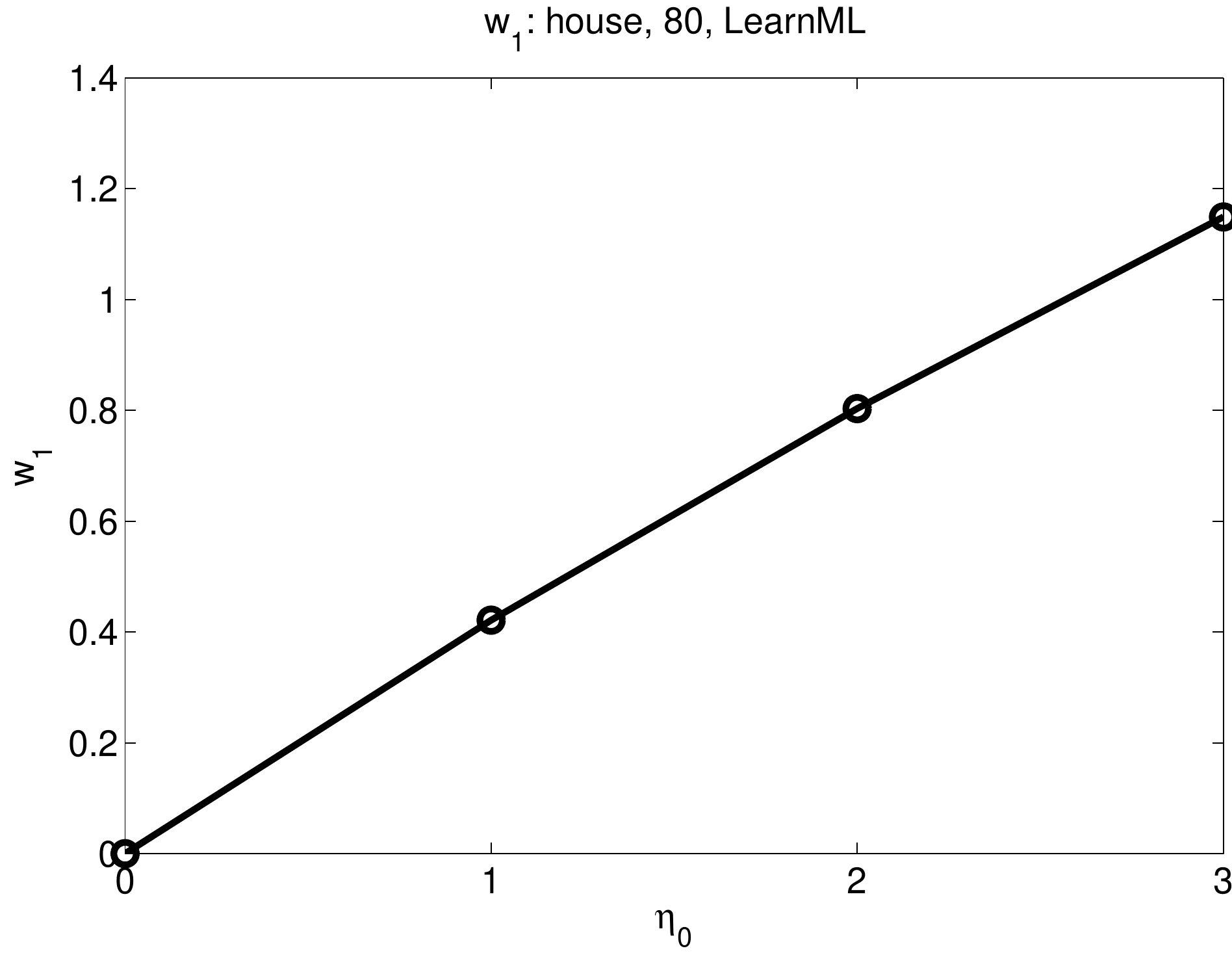}}
} \qquad
\parbox{1.25in}{\centerline{\includegraphics[width=1.25in,height=1.25in]{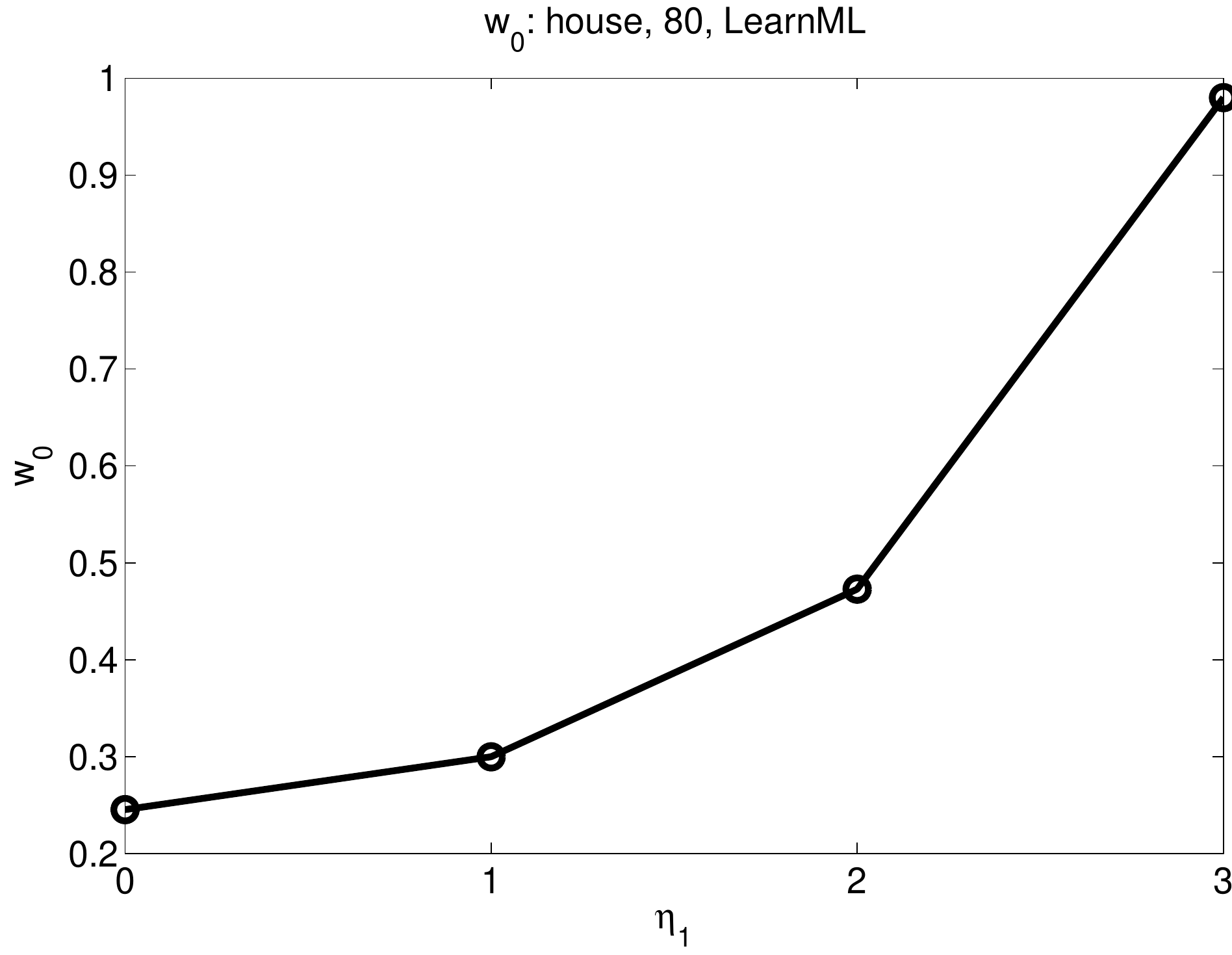}} }
\caption{\small Parameters learned by ML method, left: $w_1$, right
: $w_0$.} \label{F:LEARNEDGC}
\vspace{-0.4cm}
\end{center}
\end{figure}

In Figure~\ref{F:LEARNEDGC} , we show the discrete $g_c$ learned by
the ML algorithm for $c= 0, 1$. As expected, the learned penalty
functions resembles strongly to smooth concave ($w_1$, left in
Figure~\ref{F:LEARNEDGC}) and convex ($w_0$,right in
Figure~\ref{F:LEARNEDGC}) functions. The forms of $g_c$ functions
also provides some insight about the subsets generated for matching.
A convex penalty imposes \lq lenient\rq~  penalties on lower values
of $\eta_1$, number of label variables assuming the opposite class,
class $1$. This penalty function would be effective when there are
many subsets comprising very few (e.g., one) correct matches. For
these subsets, a convex $g_0$ would allow to let few datapoints
within the subset to assume the opposite label $1$. Examining the
matching triangles used for matching, one can verify that there are
indeed many subsets that contains one correct matches and two
incorrect matches in them. On the other hand, the triangles with all
correct matches are rare and therefore the penalty function is \lq
strict\rq~ (i.e., concave) on the value of $\eta_0$.

More plots of such learned penalty functions, as well as
non-discretized belief values (i.e., the soft assignment vector)
generated by inference algorithm and some qualitative matching
results are presented as supplementary material.


\begin{figure*}[t]
\vspace{-0.6cm}
\begin{center}
\parbox{1.3in}{\centerline{\includegraphics[width=1.3in,height=1.1in]{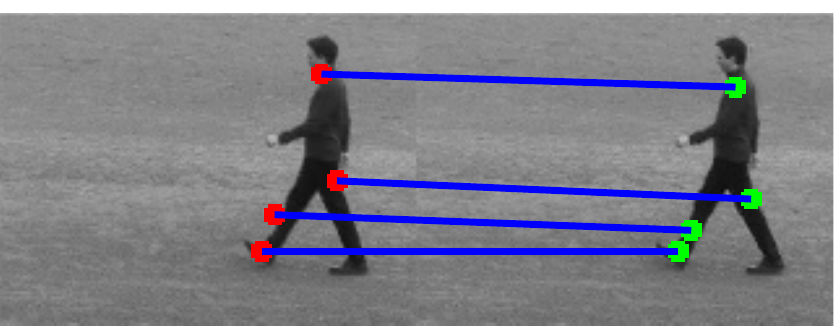}}}
\parbox{1.3in}{\centerline{\includegraphics[width=1.3in,height=1.1in]{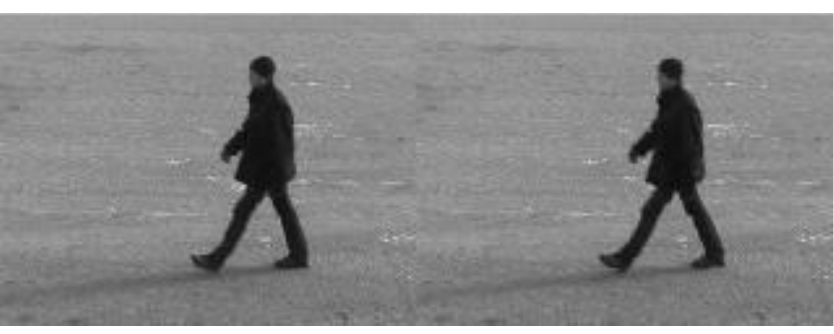}}}
\parbox{1.3in}{\centerline{\includegraphics[width=1.3in,height=1.1in]{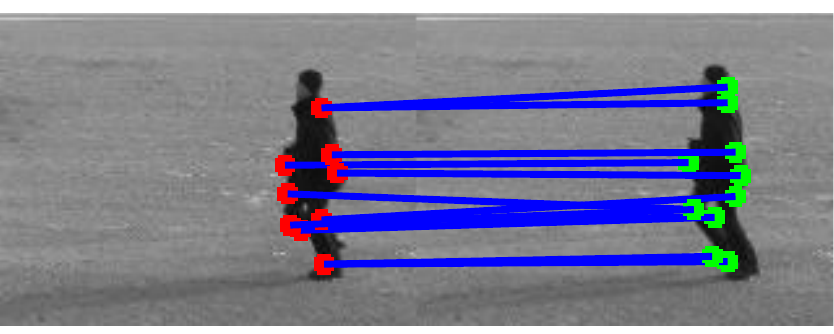}} }
\parbox{1.3in}{\centerline{\includegraphics[width=1.3in,height=1.1in]{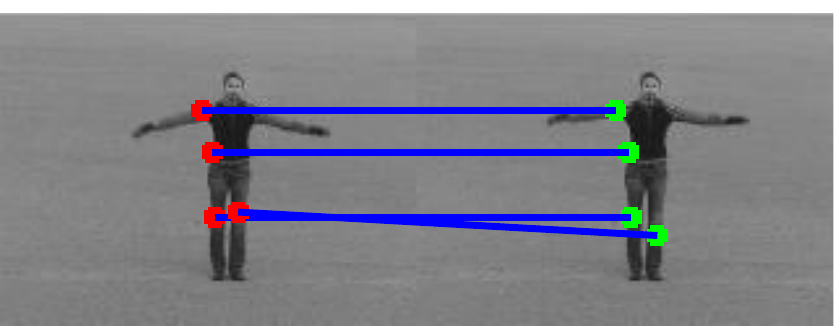}} }
\parbox{1.3in}{\centerline{\includegraphics[width=1.3in,height=1.1in]{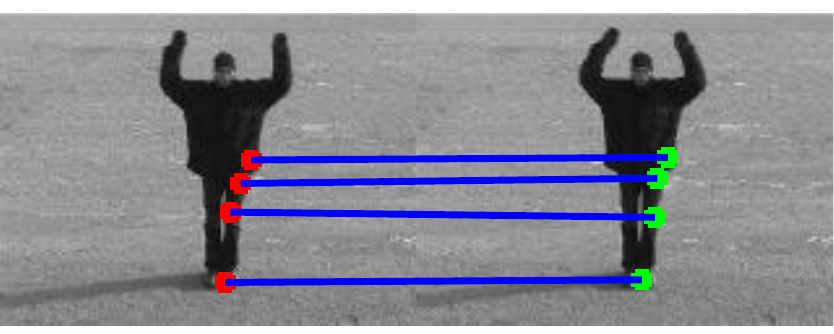}} }
\parbox{1.3in}{\centerline{\includegraphics[width=1.3in,height=1.1in]{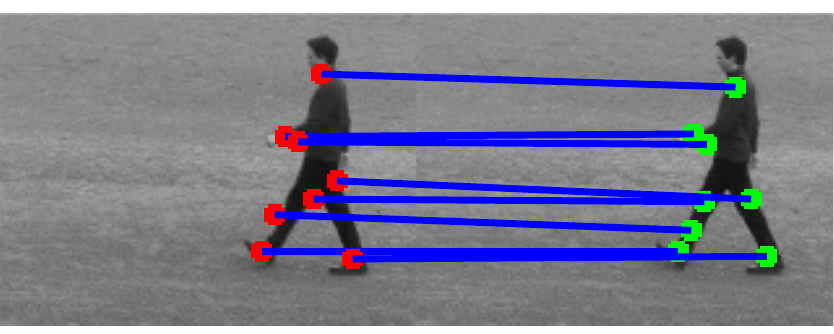}}}
\parbox{1.3in}{\centerline{\includegraphics[width=1.3in,height=1.1in]{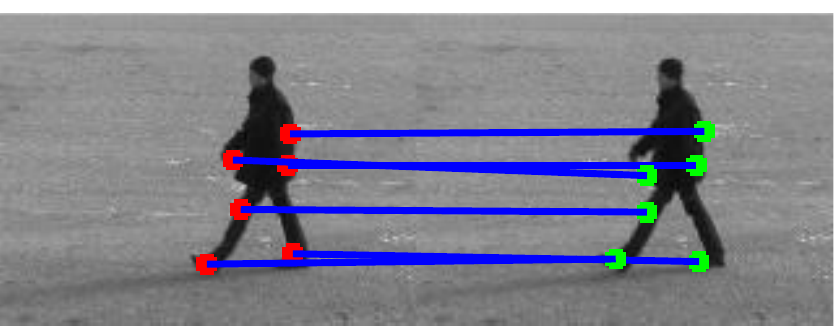}}}
\parbox{1.3in}{\centerline{\includegraphics[width=1.3in,height=1.1in]{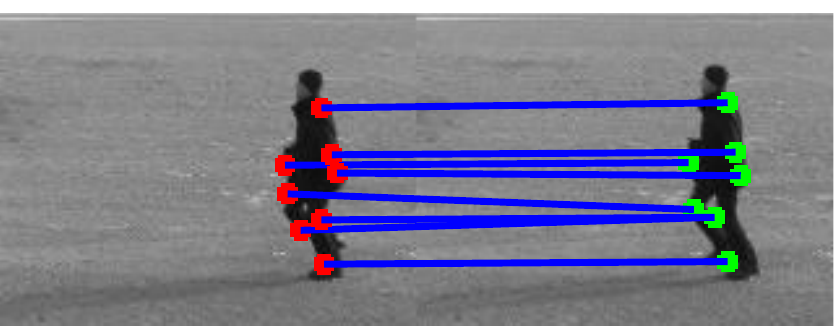}} }
\parbox{1.3in}{\centerline{\includegraphics[width=1.3in,height=1.1in]{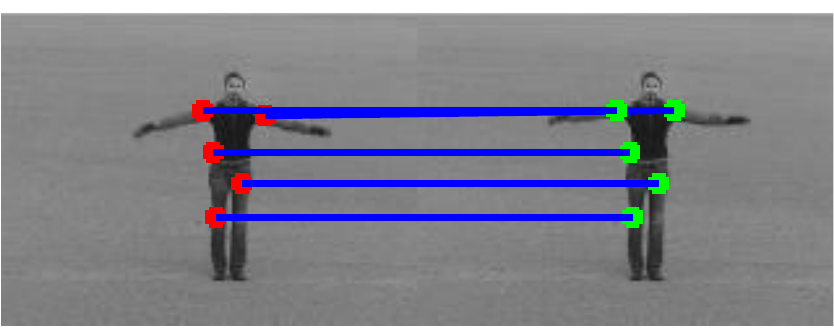}} }
\parbox{1.3in}{\centerline{\includegraphics[width=1.3in,height=1.1in]{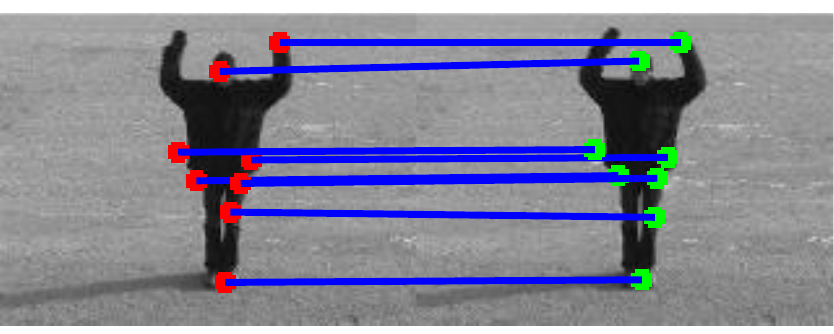}} }
\caption{\small Improvement achieved by learning. Top: results of
~\cite{parag11} using a predefined linear penalty, bottom: matches
after learning. More correct correspondences are recovered by
learned penalty function.} \label{F:WALK} \vspace{-0.5cm}
\end{center}
\end{figure*}

\begin{table*}[t]
\vspace{-0.15cm}
\begin{center}
\small \begin{tabular}{|c|c|c|c|c|c|c|c|c|c|c|c|c|}
  \hline
  method &
           \multicolumn{2}{|c|} {Jog1} &
           \multicolumn{2}{|c|} {Jog4} &
           \multicolumn{2}{|c|} {Walk1} &
           \multicolumn{2}{|c|} {Walk4} &
           \multicolumn{2}{|c|} {Wave4} &
           \multicolumn{2}{|c|} {Wave7} \\
           & True & False &
           True & False &
           True & False &
           True & False &
           True & False &
           True & False \\
\hline
Linear $g_c$&
            4.33 & 0.83 &
            5.5&1.83&
            4.89 & 0.78 &
            3.86 & 1.43&
            5& 0.5&
            6.67&2\\
\hline
Learned $g_c$ &
            4.83 & 0.67 &
            6  & 1.5&
            6.89 & 0.89 &
            5.71 & 1 &
            7.5 & 0.67&
            7.17& 1.33\\
\hline
\end{tabular}
\caption{\small Average number of correct and incorrect matches
(\emph{NOT} percentages)
 found on the image pairs. The proposed algorithm with learned penalty functions
consistently produces more true positives with less false positives
on all the sequences.} \label{T:KTHRESULT}\vspace{-0.5cm}
\end{center}
\end{table*}

\subsection{KTH Activity}

We applied our method on some KTH activity recognition
data~\cite{cateno09}. For this dataset, we chose three activities,
walking, jogging and hand waving and for each of these activities we
randomly  selected two sequences. The experimental setup is almost
same as above \emph{except} the features are detected using
Kadir-Brady (KB) keypoint detector algorithm~\cite{kadir-01keypoint}
on both the images, i.e., we do not manually select keypoints on
image. For each keypoint selected by the feature detector (KB) on
the left image, the goal is to find its best match on the right
image.

One of the objectives of this experiment is to show the necessity of
learning the penalty function instead of employing  predefined
(linear) ones. We applied the labeling algorithm with predefined
linear penalty functions and compared the results to show the
improvement achieved by learning $g_c$. For the learning algorithms,
discrete $g_c$ functions are learned using the ML estimation
procedure as before. All parameters for both methods are the same
for all the experiments in this section. Sample output matches are
shown in Figure~\ref{F:WALK}. The top row shows the output produced
by the proposed method using linear penalties, and the bottom row
shows the results produced by discrete $g_c$ trained from data. The
matching algorithm with learned penalty function were able to
extract more accurate matches than that with linear penalties.

Table~\ref{T:KTHRESULT} summarizes the quantitative matching
performances of these two methods. The results clearly show that
hypergraph labeling with learned penalty function consistently
produces better results than the same method with predefined linear
penalties. It is worth mentioning here that the proposed matching
algorithm was applied to the (spatially clustered) keypoint
locations detected by the KB detector leading to variable number of
feature locations in different images.  We manually counted the
number of correct and incorrect matches from the output for
quantitative performance evaluations.

The learned penalty functions for each of these datasets resemble
closely to those shown in Figure~\ref{F:LEARNEDGC}, please refer to
the supplementary material specific plots. These learned optimal
penalty functions are clearly non-linear which explains why
predefined linear penalty functions produce inferior matching
results.


\subsection{Caltech Aeroplane and Motorbike}

\begin{figure*}[t]
\vspace{-0.6cm}
\begin{center}
\parbox{3in}{\centerline{\includegraphics[width=3in,height=1.2in]{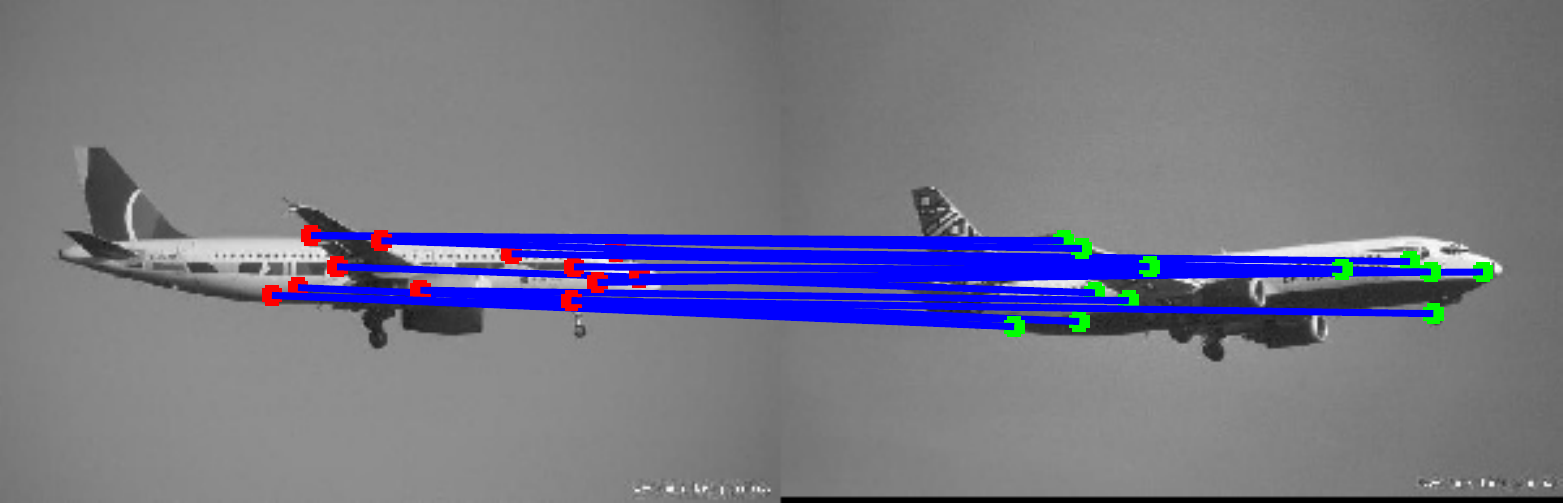}}}
\parbox{3in}{\centerline{\includegraphics[width=3in,height=1.2in]{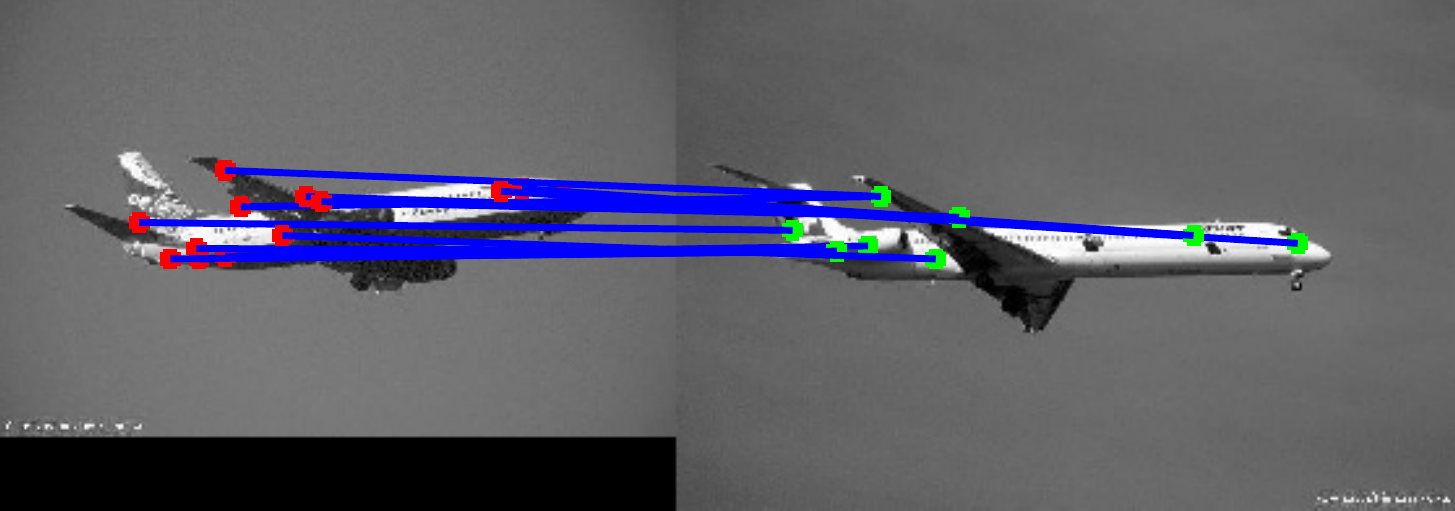}}}
\parbox{3in}{\centerline{\includegraphics[width=3in,height=1.2in]{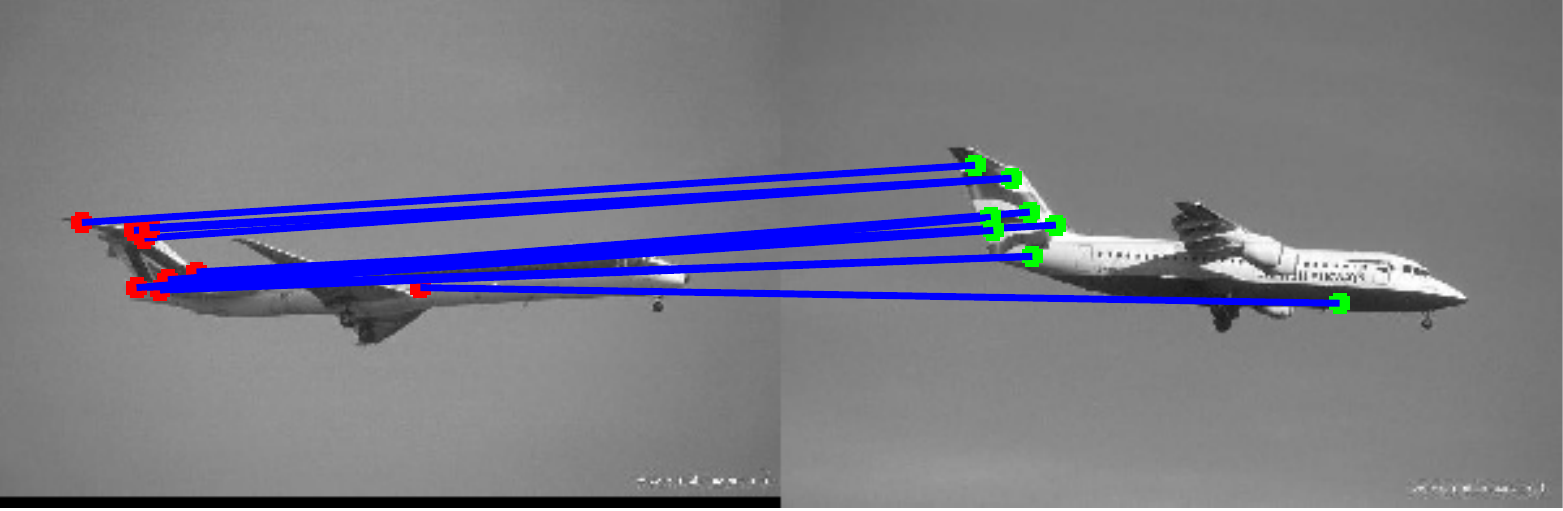}}}
\parbox{3in}{\centerline{\includegraphics[width=3in,height=1.2in]{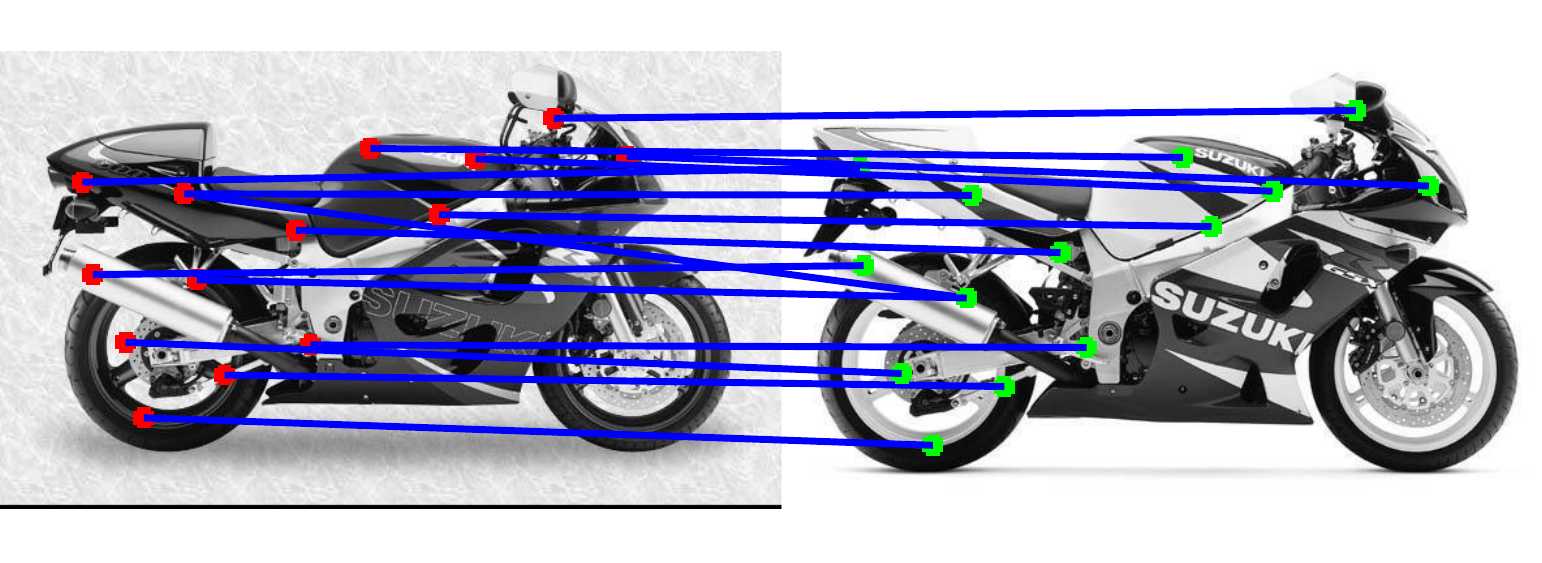}}}
\parbox{3in}{\centerline{\includegraphics[width=3in,height=1.2in]{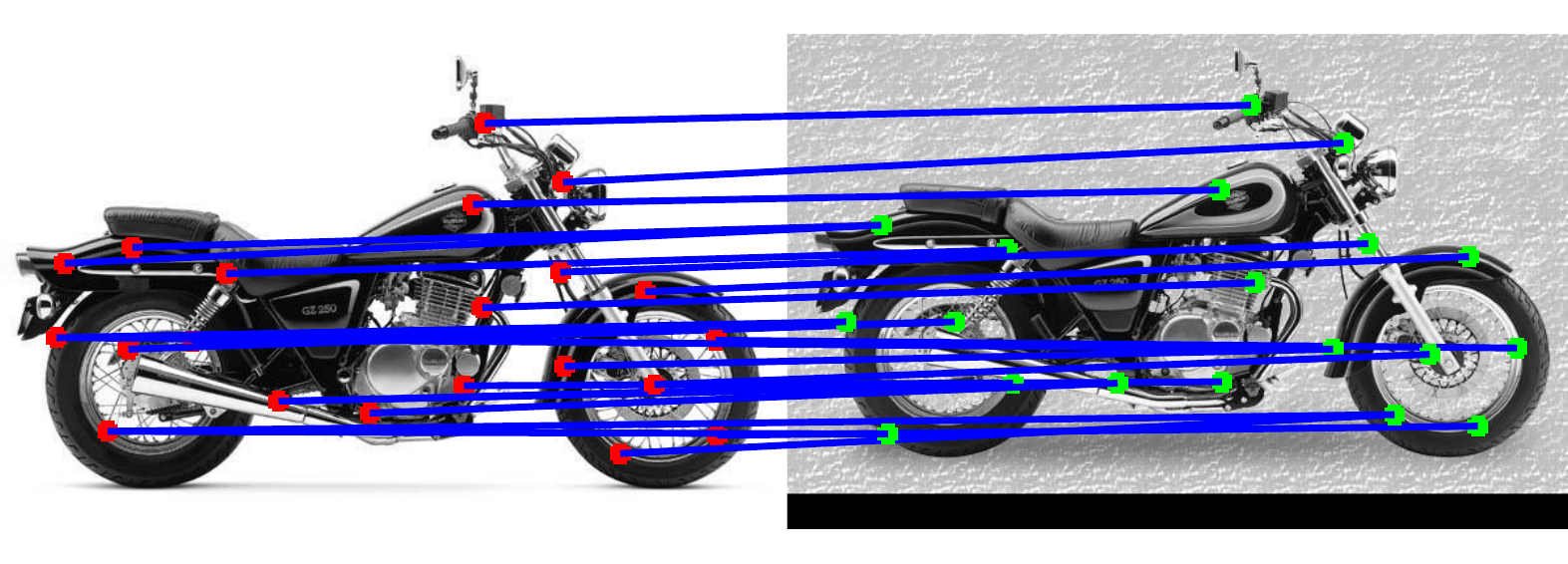}}}
\parbox{3in}{\centerline{\includegraphics[width=3in,height=1.2in]{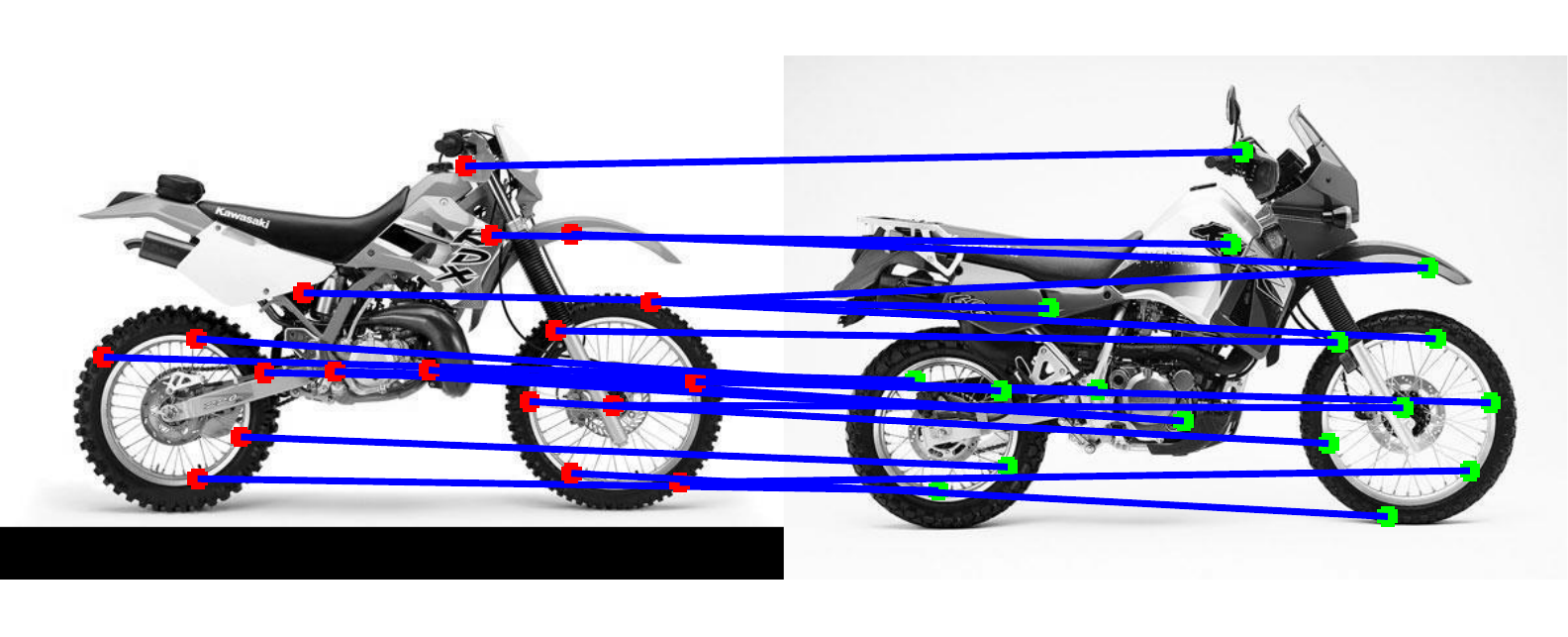}}}
\caption{\small Qualitative results on  Caltech aeroplanes and
motorbikes.} \label{F:CALTECH1} \vspace{-0.5cm}
\end{center}
\end{figure*}

Finally, we are showing some more qualitative results on Caltech
objects, such as airplanes and motorbikes, in
Figure~\ref{F:CALTECH1}. The experimental setup is exactly same as
that described in the last section. Notice that, in this experiment,
we are establishing correspondences between two different instances
of same object category, unlike the experiments described before.

\section{Discussion}\label{S:CONCL}
In this paper, we propose a novel feature matching algorithm based
on higher order information among them. The feature correspondence
problem is formulated as a hypergraph node labeling problem. A
recent algorithm that models the higher order interaction among the
datapoints using a Markov network is applied to address the labeling
problem. We describe how the associated cost function can be learned
from labeled data using existing graphical model training algorithm.
The results show that learning the cost function makes the proposed
matching algorithm more robust than other pairwise and higher order
methods.

This paper presents methods to learn the appropriate cost functions
(in terms of the penalty functions) of a hypergraph node labeling
algorithm~\cite{parag11}. Feature correspondence is one significant
application of the supervised hypergraph labeling algorithm, but the
learning procedure can benefit any applications of it. We strongly
believe learning penalty functions will improve the performances of
model estimation and object localization demonstrated
in~\cite{parag11}.

Hypergraph labeling method could potentially be applied to other
problems where learning cost functions could be advantageous. One
such problem is object boundary detection or image segmentation. We
performed a small experiment on natural images of Berkeley dataset.
The description of the procedure and sample results are shown in the
supplementary material to avoid confusion. These results suggest the
method can be used for segmentation problems, at least for specific
domain if not for natural images, with appropriately chosen image
features and model.



{ \small
\bibliographystyle{ieee}
\bibliography{iccv11matching}
}

\end{document}